\documentclass[11pt,a4paper]{article}
\usepackage[hyperref]{emnlp-ijcnlp-2019}
\usepackage{times}
\usepackage{latexsym}
\usepackage{graphicx}

\usepackage{url}
\usepackage{color,soul}
\usepackage{float}
\usepackage{amsmath}
\usepackage{amsfonts}
\usepackage{bm}
\usepackage{threeparttable}
\usepackage{algorithm}
\usepackage[noend]{algpseudocode}
\usepackage{multicol}
\usepackage{multirow}
\usepackage{booktabs}

\aclfinalcopy 

\title{Look-up and Adapt: A One-shot Semantic Parser}

\author{Zhichu Lu\thanks{~~~These two authors contributed equally} 
\\
  Carnegie Mellon University \\
  Pittsburgh, PA \\
  {\tt zhichul@cs.cmu.edu} \\\And
  Forough Arabshahi\footnotemark[1]\\
  Carnegie Mellon University \\
  Pittsburgh, PA \\
  {\tt farabsha@cs.cmu.edu}\\\And
    Igor Labutov \\
    LAER AI, Inc. \\
    New York, NY\\
  {\tt igor.labutov@laer.ai}\\\AND
    Tom Mitchell \\
  Carnegie Mellon University \\
  Pittsburgh, PA \\
  {\tt tom.mitchell@cs.cmu.edu} \\}

\date{}

\begin{document}
\maketitle
\begin{abstract}
  Computing devices have recently become capable of interacting with their end users via natural language. However, they can only operate within a limited ``supported'' domain of discourse and fail drastically when faced with an out-of-domain utterance, mainly due to the limitations of their semantic parser. In this paper, we propose a semantic parser that generalizes to out-of-domain examples by learning a general strategy for parsing an \emph{unseen} utterance through adapting the logical forms of \emph{seen} utterances, instead of learning to generate a logical form from scratch. Our parser maintains a memory consisting of a representative subset of the seen utterances paired with their logical forms. Given an unseen utterance, our parser works by looking up a similar utterance from the memory and adapting its logical form until it fits the unseen utterance. Moreover, we present a data generation strategy for constructing utterance-logical form pairs from different domains. Our results show an improvement of up to 68.8\% on one-shot parsing under two different evaluation settings compared to the baselines.
\end{abstract}

\section{Introduction and Background}
Speech recognition technologies are achieving human parity \cite{speechrecognition}. As a result, end users can now access different functionalities of their phones and computers through spoken instructions via a natural language processing interface referred to as a conversational agent. Current commercial conversational agents such as Siri, Alexa or Google Assistant come with a fixed set of simple functions like setting alarms and making reminders, but are often not able to cater to the specific phrasing of a user or the specific action a user needs. However, it has recently been shown that it is possible to add new functionalities to an agent through natural language \emph{instruction} \cite{LIA, LIAv2}.

For example, assume that the user wants to add a functionality for resetting an alarm based on the weather forecast for the next day, as demonstrated by the following utterance: ``whenever it snows at night, wake me up 30 minutes early''. The user can instruct this task to the agent by breaking it down into a sequence of actions that the agent already knows.
\vspace{-0.5em}
\begin{itemize}
    \setlength\itemsep{0.5 mm}
    \item check the weather app, 
    \item see if the forecast is calling for snow,
    \item if yes, then reset the time of the alarm to 30 minutes earlier.
\end{itemize}
\vspace{-0.5em}
This set of instructions result in a logical form or a semantic parse for this specific new utterance. However, this approach can be used in practice only if the agent is capable of generalizing from this single new utterance to similar utterances such as ``if the weather is rainy, then set an alarm for 1 hour later''. We refer to this problem as \emph{one-shot semantic parsing}.

In this paper, we address this one-shot semantic parsing task and present a semantic parser that generalizes to out-of-domain utterances by seeing a single example from that domain. While state of the art neural semantic parsers are flexible to language variations, they need plenty of examples from the new domain to be able to parse an utterance from that domain, which is not possible in our scenario. On the other hand, grammar parsers are not robust to the flexibility of language because of their use of string matching. Therefore, we propose a method that preserves the robustness of neural semantic parsers while addressing the data sparsity of the one-shot semantic parsing task. We present a general strategy for ``adapting'' logical forms rather than constructing them from scratch by ``looking up'' similar sentences that we know how to parse and changing their logical forms until they fit the new utterances. These logical forms are looked-up from a memory that contains a representative subset of previously seen utterance-logical form pairs. Once this general strategy is learned, the parser can be extended to parse an utterance from a new domain by adding one new example of that domain to the memory. 

We propose a dataset generation method that allows us to evaluate the effectiveness of our model in a one-shot setting. We show that we generate reasonably good utterances while creating different experimental setups and scenarios for evaluation. 

\paragraph{Summary of Results} In this paper we propose a novel neural semantic parser for the task of one-shot semantic parsing. We design two different experiments to evaluate the parsing accuracy. We show that our approach improves the performance of neural semantic parsers by a significant margin across 6 different domains of discourse. Moreover, we present a detailed analysis of our proposed model and the performance of its different components through an oracle study.

\section{Problem Definition}
\label{definition}
In this section, we introduce the basic terminology used in the paper and formally define the task of one-shot semantic parsing.

A semantic parser takes as input an \emph{utterance} and outputs a corresponding \emph{logical form}. An utterance is a sequence of words and a logical form is an s-expression capturing the meaning of the utterance.

For example, ``parents of John's friends" is an utterance and \texttt{(field parent (field friend John))} is its logical form. 

We use a synchronous context free grammar (SCFG), which is a generalization of the context-free grammar (CFG), to generate grammar rules \cite{SCFG}. A rule in an SCFG has a left hand side, which is a non-terminal category, and two right hand sides, referred to as the source and the target. For our semantic parsing task, the source corresponds to an utterance, and the target corresponds to a logical form. We denote a grammar rule with the small letter $g$. A set of grammar rules, denoted by the capital letter $G$, span a domain of utterances. In this paper, we define $domain(G)$ as the set of all \textit{utterance}-\textit{logical form} pairs generated by a set of grammar rules $G$.
A sample SCFG $G_{sample}$ with its domain are provided in Tables~\ref{tab:example_scfg_person} of Appendix~\ref{app:scfg} and Table~\ref{tab:example_domain} below, respectively.

\begin{table}[h]
\centering \small
\begin{tabular}{|ll|}
\hline \hline
Utterance & Logical Form\\
\hline \hline
John & \texttt{john}\\
\hline
Mary & \texttt{mary}\\
\hline
 parents & \texttt{parent}\\
\hline
 children & \texttt{child}\\
\hline
 parents of John & \texttt{(field parent john)}\\
\hline
 parents of Mary & \texttt{(field parent mary)}\\
\hline
 children of John & \texttt{(field child john)}\\
\hline
 children of Mary & \texttt{(field child mary)}\\
\hline
 John 's parents& \texttt{(field parent john)}\\
\hline
 Mary 's parents & \texttt{(field parent mary)}\\
\hline
 John 's children& \texttt{(field child john)}\\
\hline
 Mary 's children & \texttt{(field child mary)}\\
\hline
\hline
\end{tabular}
\caption{Domain of the sample SCFG $G_{sample}$. The rules of this SCFG are presented in the Appendix \ref{sec:rules}. The domain of the sample SCFG is the set of all $\langle$source, \texttt{target}$\rangle$ pairs listed in this table.}
\label{tab:example_domain}
\end{table}
In \textbf{one-shot semantic parsing} we are given as input a subset of utterance and logical form pairs from $domain(G)$ and a single utterance and logical form pair from $domain(G')$. The goal of one-shot semantic parsing is to parse utterances from $domain(G')$ that are not in the input.

Let us provide an example to make this more clear. Consider that we are given utterance and logical form pairs from $domain(G_{\text{sample}})$ in Table~\ref{tab:example_domain} as well as the following utterance and logical form pair as input.
\begin{align*}
    \langle \text{friends of John}, \texttt{(field friend john)}\rangle
\end{align*}
The goal of one-shot semantic parsing is to parse examples such as
\begin{align}
    &\langle \text{friends of Mary}, \\ \nonumber
    & \hspace{0.5em}\texttt{(field friend Mary)}\rangle \\
    &\langle \text{parents of Mary's friends}, \\ \nonumber
    & \hspace{0.5em} \texttt{(field parent\!(field friend Mary)}\rangle 
\end{align}
which are not in $domain(G_{\text{sample}})$.

This is a challenging task for a complex grammar since the domain of a typical grammar consists of hundreds of thousand of examples. Therefore, being able to parse all variations of utterances in the domain by seeing only a single example from it does not have a trivial solution. Grammar-based semantic parsers usually rely on string matching which limits their robustness to natural language variation. Neural semantic parsers are more robust compared to grammar-based parsers. However, their performance significantly drops in such a data-hungry setting. In the next section, we present our look-up adapt semantic parser that addresses the challenges of neural semantic parsers in a data-sparse scenario. 

\section{Look-up and Adapt}
In this section, we propose a novel neural network architecture for one-shot semantic parsing. We are given a set of utterance-logical form pairs as input and our goal is to output a semantic parse for an unseen query utterance. Our model is able to generate a logical form for the utterance by ``looking up'' a similar utterance from a pool of utterance-logical form pairs and ``adapting'' its logical form. This pool of utterances is maintained in a ``memory'' and consist of a representative subset of the input utterance-logical form pairs. We will show that in the data-sparse scenario of one-shot semantic parsing, adapting known logical forms is easier compared to generating a logical form from scratch. We will also discuss that it is important what subset of the data is included in the memory.

\begin{figure*}
    \centering
    \includegraphics[width=0.85\textwidth]{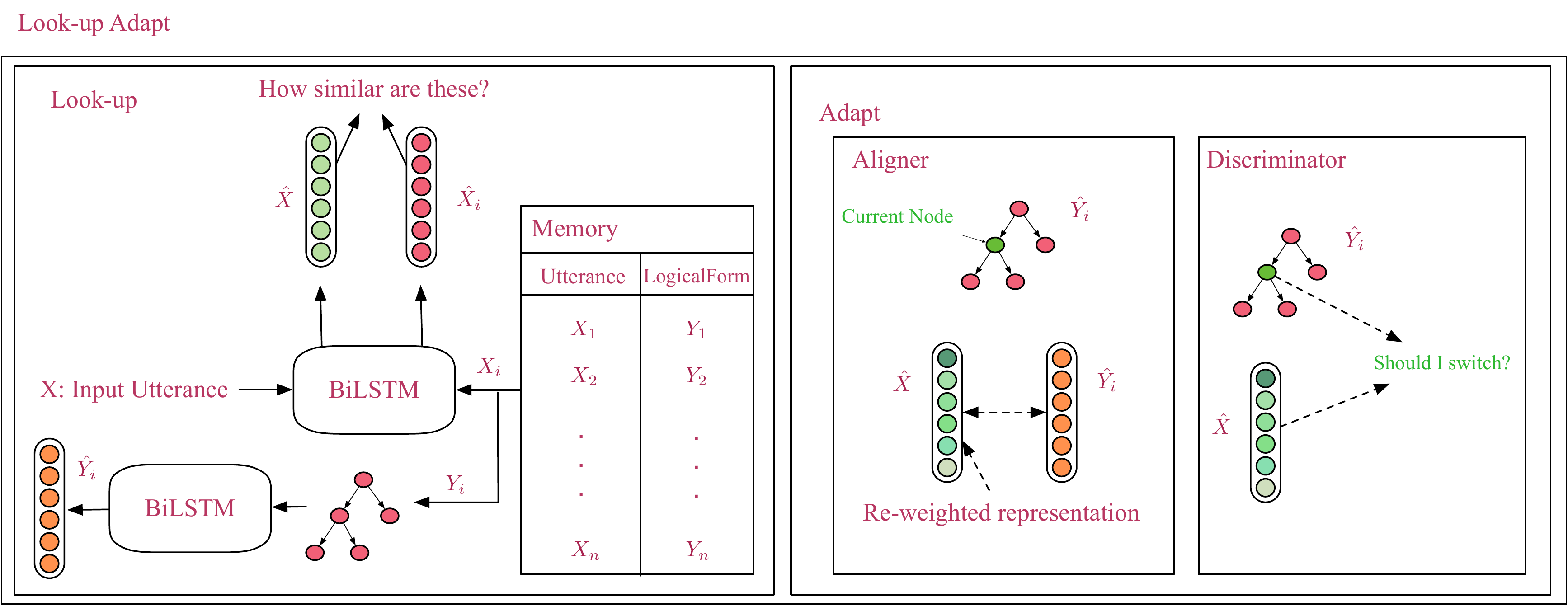}
    \caption{Look-up Adapt semantic parser. Our model consists of two main modules, namely ``Look-up'' (left box) and ``Adapt'' (right box). The look-up module is responsible for retrieving utterance-logical form pairs from the memory that are similar to an input query utterance $X$. The `Adapt'' module is further broken down into two modules ``Aligner'' and ``Discriminator''. Given a looked-up utterance and logical form pair, the aligner module traverses the logical form tree and updates the utterance representation accordingly while the discriminator module decides weather to switch the current node with another one from the memory or keep it as is.}
    \label{fig:model}
\end{figure*}

Our model consists of two main modules, namely look-up \S~\ref{sec:lookup} and adapt \S~\ref{sec:adapt}. Figure \ref{fig:model} shows a sketch of our proposed model and the main look-up adapt algorithm is given in Algorithm~\ref{alg:lookup_adapt}. The look-up module is responsible for retrieving utterance-logical form pairs from the memory, using two Bidirectional LSTM encoders. The adapt module is responsible for adapting the retrieved logical form until it results in the correct semantic parse. The adapt module has two sub-modules, namely the aligner and the discriminator. The aligner is responsible for aligning the logical form with the query utterance and the discriminator decides which parts of the logical form should be swapped with a new one from the memory. In the following sections, we will define each sub-module of our proposed model in detail and propose a loss for training the model in an end-to-end fashion.

Before describing the model components, let us start by introducing the notation we will use in the following sections. Bold capital letters represent matrices, and bold lower case letters represent vectors. $\bm{X}$ denotes a distributed representation of an utterance, where each column of $\bm{X}$ is the representation for a word in the utterance. $\bm{Y}$ denotes a distributed representation of a logical form, where each column of $\bm{Y}$ is the representation for a predicate in the logical form. \texttt{T} is a tree representation of the logical form. $\bm{w}$ and $\bm{w}'$ denote attention weights that sum to at least 0 and at most 1.

\begin{algorithm}
\small
\caption{LookUpAndAdapt}
\label{alg:lookup_adapt}
\begin{algorithmic}[1]
\Function{LookUpAndAdapt}{$\texttt{Memory}, \bm{X}', \bm{w}'$}
  \State $(\bm{X}, \bm{Y}, \texttt{T}) \gets \Call{LookUp}{\texttt{Memory}, \bm{X}', \bm{w}'}$
  \ForAll{$\texttt{t}$ \textbf{in} $\texttt{T.children}$ } 
    \State $\texttt{t}' \gets \Call{Adapt}{\texttt{Memory}, \bm{X}', \bm{Y}, \texttt{t}, \bm{w}'}$
    \State $\texttt{T.children.replace(t, t')}$
  \EndFor
  \State \Return{$\texttt{T}$}
\EndFunction
\end{algorithmic}
\end{algorithm}

\subsection{Look-up}
\label{sec:lookup}
In this section, we describe our look-up strategy and discuss its sub-modules in detail. 

\paragraph{Encoder}
Bidirectional RNNs have been successfully used to represent sentences in many areas of natural language processing such as question answering, neural machine translation, and semantic parsing \cite{nmt, mc, nerualsemanticparsing}. We use Bidirectional LSTMs where the forward LSTM reads the input in the utterance word order ($x_1, x_2, ..., x_T$) and the backward LSTM reads the input in reversed order ($x_T, x_{T-1}, ..., x_1$). The output for each word $x_i$ is the concatenation of the $i$th hidden state $h^{fwd}_i$ of the forward LSTM and the $T-i$th hidden state $h^{bck}_{T-i}$ of the backward LSTM. As shown in Figure~\ref{fig:model}, we use two Bidirectional LSTMs, one to encode the utterance, and one to encode the logical form. The utterance is treated as a sequence of words, where each word is represented as a pretrained GloVe \cite{pennington2014glove} embedding. The logical form is treated as a sequence of predicates and parentheses, each of which is also represented as a pretrained GloVe embedding. The motivation to use GloVe embeddings for both words and predicates is to avoid the problem of unknown words/predicates encountered in the one-shot utterance's domain.

\paragraph{Retrieving from the memory}
The memory consists of a subset of the utterance-logical form pairs given as input to the algorithm. We refer to the number of pairs in the memory as its size and denote it with $n$. We would like to note that the algorithm's success depends on having a memory that captures the structure of the domain. In this paper, we make a simplifying assumption that the memory is given by an oracle, and it consists of one example per grammar rule. Other choices for the memory are also possible but exploring them falls out of the scope of this paper.  An interested reader is referred to Appendix \ref{app:memory} for a more detailed discussion. In the future we wish to automate the acquisition of this memory. 

Given an encoded query utterance $\bm{X}'$, we model the retrieval as a classification over examples in the memory. 
We compute a fixed size query vector $\bm{x}' = \bm{X}'\cdot \bm{w}'$ which is a weighted average of the column vectors in $\bm{X}'$. We compute a fixed-size key vector for every example in the memory $\bm{x} = \sum_{i=1}^T \bm{X}_i$, as the mean of the column vectors of $\bm{X}$.

Given a query vector $\bm{x}'$ and vectors $\bm{x}_i$ from the memory for $i = 1,2,...,n$ where $n$ is the memory size, we model the probability of retrieving the $i$th entry from memory using equation~\ref{eqn:retrieve}
\begin{equation}
    P(i|\bm{x}') = \frac{\exp (e_i)}{\sum_{j=1}^n \exp(e_j)} \label{eqn:retrieve}
\end{equation}
where 
\begin{equation}
    e_i = f([\bm{x}', \bm{x}_i, \bm{x}' \odot \bm{x}_i, |\bm{x}' - \bm{x}_i|, \bm{x'}^TW_1\bm{x}_i])
\end{equation}
where $\odot$ is element-wise product,$[a,b]$ is concatenation of vectors $a$ and $b$, and $f$ is a Multi-Layer-Perceptron with a single hidden layer and ReLU activation function. This way of featurizing $\bm{x}'$ and $\bm{x}$ was adapted from \cite{gated}. \textsc{LookUp} greedily returns the example with the highest probability of being selected.
\subsection{Adapt}
\label{sec:adapt}
The algorithm for \textsc{Adapt} is given in Algorithm~\ref{alg:adapt}. The adaptation is two phase. First an alignment from \texttt{T} to relevant words in the new utterance $\bm{X}'$ is computed. Then a decision is made on whether this subtree fails to match the relevant words in the utterance and needs to be replaced. If yes, a recursive call to \textsc{LookUpAndAdapt} will be made with an updated attention $\bm{w}$ to focus on words relevant to the subtree, and the returned value will be propagated and eventually used as a replacement for the current subtree. Otherwise, recursive calls to \textsc{Adapt} will be made to check the children of the current subtree. For an example see Table~\ref{tab:trace}.\par  
The input \texttt{Memory} to the algorithm is used in case a recursive call to LookUpAndAdapt is needed. $\bm{X}'$ is the representation of the new utterance. $\bm{Y}$ is the representation of the current logical form. \texttt{T} is a subtree of $\bm{Y}$ that we wish to adapt. $\bm{w}'$ is the attention weights of the parent of \texttt{T}, to be used as a mask when the alignment of \texttt{T} is computed. In the following space we describe the aligner module for alignment and the discriminator module for scoring a match between a subtree and part of the new utterance.
\begin{algorithm}
\small
\caption{Adapt}
\begin{algorithmic}[1]
\Function{Adapt}{$\texttt{Memory}, \bm{X}', \bm{Y}, \texttt{T}, \bm{w}'$}
 \State \textbf{Phase 1: aligning \texttt{T} to relevant words in $\bm{X}'$}
  \State $\bm{w} \gets \Call{Align}{\bm{X}', \bm{Y}, \texttt{T}, \bm{w}'}$
  \State $\bm{x}' \gets \bm{X}' \cdot \bm{w}$
  \If{$\sum_{i} \bm{w}_i > 1$}
  \State $\bm{x}' \gets \bm{x}' \div \sum_{i} \bm{w}_i$
  \EndIf
  \State
   \State \textbf{Phase 2: deciding whether to replace \texttt{T} and then recurse }
  \State $P(fail\ to\ match) \gets \Call{Discriminate}{\bm{x}', \bm{Y}, \texttt{T}}$
  \If{$P(fail\ to \ match) > 0.5$} 
  \State \Return $\Call{LookUpAndAdapt}{\texttt{Memory}, \bm{X}', \bm{w}}$ 
  \Else
    \ForAll{$\texttt{t}$ \textbf{in} $\texttt{T.children}$ }
        \State $\texttt{t}' \gets \Call{Adapt}{\texttt{Memory}, \bm{X}', \bm{Y}, \texttt{t}, \bm{w}}$
    \State $\texttt{T.children.replace(t, t')}$
    \EndFor
    \State \Return{T}
  \EndIf
\EndFunction
\end{algorithmic}
\label{alg:adapt}
\end{algorithm}

\paragraph{Aligner Module}
The aligner module produces an attention score over the new utterance given a subtree of a logical form. Inspired by the gating attention mechanism in \cite{gated}, we model attention for every word in the utterance as the maximum attention given by any predicate in the subtree \texttt{T}.
\vspace{-1em}
\begin{equation}
    \bm{w}_i = \max_{j \in span(\texttt{T})} \sigma(s_{ij})
\end{equation}
where $\sigma$ is the sigmoid function and
\begin{align}
    s_{ij} &= g([\bm{X}_i, \bm{Y}_j, \bm{Y}_p, \bm{X}_i \odot \bm{Y}_j, \bm{X}_i \odot \bm{Y}_p, \\ \nonumber
    & |\bm{X}_i - \bm{Y}_j|, |\bm{X}_i - \bm{Y}_p|, \bm{X}_i^T W_2 \bm{Y}_j, \bm{X}_i^T W_p \bm{Y}_p])
\end{align}
where $\odot$ is element-wise product, $[a,b]$ is concatenation of vectors $a$ and $b$, $\bm{X}_i$ is the $i$th column of $\bm{X}$, $\bm{Y}_i$ is the $i$th column of $\bm{Y}$, and $g$ is a learnable linear transformation. $\bm{Y}_p$ is the representation of the parent predicate of the subtree \texttt{T} (e.g. the parent predicate is \texttt{field} for subtree \texttt{john} of logical form \texttt{(field parent john)}). We found that adding this feature helps the model learn to align arguments of predicates better. \par 
We also found that constraining the attention further by requiring that the attention of a child node $\bm{w}$ be a refinement of the attention of a parent node $\bm{w}'$ facilitates good alignment. This is modeled by the update rule 
\begin{equation}
    \bm{w}_i \gets \bm{w}_i \cdot \sqrt{\bm{w}'_i}
\end{equation}
Observe that if the attention score of the parent node is very low for some word, the child is not going to be able to look at it. This encourages the parent node to attend to words not just relevant to itself but also the children, and encourages the child to refine rather than drastically change from the alignment of the parent node. \par
Given the output weights of the aligner module, we compute a fixed-size representation for the relevant words in the utterance to the current subtree \texttt{T} by first taking the weighted average $\bm{x}' = \bm{X}'\cdot \bm{w}$ where $\cdot$ is matrix product. If $\sum_{i} \bm{w}'_i > 1$, we will divide $\bm{x}'$ by $\sum_{i} \bm{w}'_i$ to normalize it. $\bm{x}'$ is then compared against the representation of the subtree by the discriminator to produce a confidence score indicating whether the subtree matches the words it aligns to in the utterance.
\paragraph{Discriminator Module}
The discriminator module learns to tell when a subtree fails to match the meaning of the words in its alignment. Given a subtree \texttt{T} of a logical form $\bm{Y}$, its fixed-size representation is computed as a mean of the representations for each predicate in the subtree using the formula $\bm{y} = \sum_{i \in span(\texttt{T})} \bm{Y}_i$, where $\bm{Y}_i$ is the $i$th column of $\bm{Y}$, and $span(\texttt{T})$ is the set of indices corresponding to all the predicates in the subtree \texttt{T}. For example, the fixed-size representation for the subtree \texttt{john} in logical form \texttt{(field parent john)} is just $\bm{Y}_4$ (indexing from 1, and also counting parentheses because they are also encoded in $\bm{Y}$).\par We model the confidence that the subtree representation $\bm{y}$ fails to match the meaning of the aligned words $\bm{x}'$ as
\begin{equation}
    P(fail\ to\ match|\bm{x}', \bm{y}) = \sigma(d) \label{eqn:adapt}
\end{equation}
where $\sigma$ is the sigmoid function and
\begin{equation}
    d = h([\bm{x}', \bm{y}, \bm{x}'\odot \bm{y}, |\bm{x}' - \bm{y}|, \bm{x}'^T W_3 \bm{y}])
\end{equation}
$\odot$ is element-wise product, $[a,b]$ is concatenation of vectors $a$ and $b$, and $h$ is a learnable linear transformation. This confidence score is used by Algorithm~\ref{alg:adapt} to decide whether to replace the subtree or to keep it.

\section{Training}
In training, we maximize the log conditional likelihood of the data. Due to our selection of memory (one example for each grammar rule in $G$), for a given $x$ there is a unique sequence of retrieval and adaptation actions that lead to the production of the correct $y$. Specifically, letting $a_i$ be the $i$th action in the correct sequence of $l$ actions, we define the probability of producing $y$ given $x$ as the probability of taking the sequence of actions $a_1,...,a_l$. We decompose this joint probability into products of conditional probabilities of taking action $i$ given the previous actions and the input $x$
\begin{equation}
P(y|x) = \prod_{i=1}^l P(a_i | a_1, a_2, ..., a_{i-1},x)
\end{equation}
If $a_i$ is a retrieval action,
\begin{equation}
 P(a_i | a_1, a_2, ..., a_{i-1},x) = P(i|\bm{x}')
\end{equation}
where $P(i|\bm{x}')$ is defined in Equation~\ref{eqn:retrieve}. If $a_i$ is an adaptation decision, i.e. to replace a particular subtree
\begin{equation}
 P(a_i | a_1, a_2, ..., a_{i-1},x) = P(fail\ to\ match|\bm{x}', \bm{y})
\end{equation}
where $P(fail\ to\ match|\bm{x}', \bm{y})$ is defined in Equation~\ref{eqn:adapt}. \par 
\section{Dataset Generation}
\label{sec:dataset_generation}
Using existing semantic parsing datasets to directly evaluate one-shot parsing is difficult, because evaluation of one-shot parsing requires grouping of utterances by structural similarity to construct domains.

Therefore, we generate our own data \footnote{Our dataset can be accessed at \url{https://github.com/zhichul/lookup-and-adapt-parser-data}.} based on a synchronous context free grammar. We evaluate our approach on the generated dataset. Although our dataset is synthetic, it has the properties needed to evaluate one-shot parsing: 
\begin{itemize}
\setlength\itemsep{0.5 mm}
    \item clear distinction of domains
    \item plenty of examples for rules in the old domain
    \item one example for each rule in the new domain 
    \item a separate evaluation set containing variations of the new rules
\end{itemize}
It is worth noting that although datasets such as OVERNIGHT \cite{wang2015building} do have clear domain distinction, they do not have the other properties needed for a good evaluation of one-shot parsing. In the future, we are planning to use crowd-sourcing to rephrase the generated utterances in order to make them more natural.

We have two topics of discourse in the SCFG that we use to generate our data. The first is accessing some field of some object, and the second is setting some field of some object to some value or some field of another object. These are general purpose instructions that can express many common intents, such as looking up the location of a restaurant, getting the phone number of a contact, and setting reminders and alarms.

We have six domains of discourse in the SCFG that we use to generate our data -- \emph{person}, \emph{restaurant}, \emph{event}, \emph{course}, \emph{animal}, and \emph{vehicle}. Table~\ref{tab:dataset_examples} includes sample sentences and their logical form for every domain of discourse.
\begin{table*}[h]
\centering 
\resizebox{1.8\columnwidth}{!}{%
\begin{tabular}{|cll|}
\hline \hline
& Domain & Utterance \& Logical Form\\
\hline \hline
$a_1$&\emph{person}& the hometown field of john \\
&& (field (relation hometown) (person john))\\
$a_2$&\emph{person}& set her 's parents with classmates \\
&& (set (field (relation parent) (person reference)) (person classmate))\\
\hline
$b_1$&\emph{restaurant}& irish restaurant instances 's price field \\
&& (field (relation price) (restaurant irish))\\
$b_2$&\emph{restaurant}& set address field of all irish restaurant as indian restaurant 's price
 \\
&& (set (field (relation address) (restaurant irish)) (field (relation price) (restaurant indian)))\\
\hline
$c_1$&\emph{event}& the start time of lectures\\
&& (field (relation start) (event lecture)) \\
$c_2$&\emph{event}& set the attendants of that event to organizers field of receptions instances
 \\
&& (set (field (relation attendant) (event reference)) (field (relation organizer) (event reception)))\\
\hline
$d_1$&\emph{course}& size of my history course instances \\
&& (field (relation size) (course history))\\
$d_2$&\emph{course}& set all history course instances 's prerequisite field as physics course
\\
&& (set (field (relation prerequisite) (course history)) (course physics))\\
\hline
$e_1$&\emph{animal}& life span of all lion instances\\
&&(field (relation span) (animal lion))\\
$e_2$&\emph{animal}& set fish instances 's family field with dog \\
&& 
(set (field (relation family) (animal fish)) (animal dog))\\
\hline
$f_1$&\emph{vehicle}& the source field of all buses
 \\
&& (field (relation source) (vehicle bus))\\
$f_2$&\emph{vehicle} & set operators of all subways as buses instances \\
&& (set (field (relation operator) (vehicle subway)) (vehicle bus))\\
\hline
\hline
\end{tabular}
}
\caption{Sample utterance-logical form pairs from each domain of discourse. Sentences $a_1$ ... $f_1$ parse to commands that retrieve a field of an object. Sentences $a_2$ ... $f_2$ parse to commands which set a field of an object to a constant or a field of another object. The grammar we use do not constrain the type of the set command. As a consequence, some sentences generated are not semantically correct according to a human interpreter but still they preserve the sentence structure. Sentence $b_2$, $e_2$, and $f_2$ are examples of this phenomena.}
\label{tab:dataset_examples}
\end{table*}

Since one-shot parsing has two phases, our dataset is slightly different from a typical dataset consisting of a single collection of \textit{utterance}-\textit{logical form} pairs. In the following space we describe the sets of data generated for each domain. In the next section we describe how we can use this dataset to design two different one-shot parsing evaluation setups. 

For each domain $d$, we define a $G_d$ and randomly generate examples from $domain(G_d)$ to construct an ``old" set of examples $D_{old,d}$. We also generate a representative subset $M_{old,d}$ to be used as the memory of our model. As described in \S~\ref{sec:lookup}, $M_{old, d}$ contains one example for each grammar rule in $G_d$. 

In addition, we define a set of new rules $G_d'$ disjoint from $G_d$ for each domain and generate one example $e_{i,d}$ for each new rule $g'_{i,d} \in G'_{d}$, and store them into the memory $M'_{new, d}$. These are the one-shot examples. The extended memory $M_{new, d} = M_{old, d} \cup M'_{new, d}$.

Finally, we generate evaluation sets $E_{old, d} \subset domain(G_d)$ and $E_{new, d} \subset domain(G_d\cup G_d') - domain(G_d)$ where $-$ denotes set difference. In our experiments we split the evaluation sets randomly into a development set and a test set of the same size.

Some statistics of our generated dataset is presented in Appendix \ref{sec:data}. In the next section, we describe how we use this data to generate two different experimental scenarios for evaluating one-shot semantic parsing.

\section{Results and Evaluation}
We evaluate our approach in six domains, namely person, restaurant, event, course, animal, and vehicle, and in two one-shot parsing scenarios, namely \emph{extension} and \emph{transfer}. In the next section we define these two different scenarios. We compare the performance to a sequence-to-sequence parser with attention which is defined in the following paragraphs. 

\paragraph{Baseline}
Our baseline is a sequence-to-sequence parser with attention. We use a 1-layer Bidirectional LSTM as the encoder for the utterance, and a 1-layer LSTM as the decoder for the logical form. We initialize the decoder with a learned projection of the last hidden state of the encoder. The inputs to the encoder are GloVe word embeddings, and the inputs to the decoder are concatenations of embeddings of logical predicates and context vectors. Context vectors are attention weighted averages of projected encoder states. The attention weights over the utterance for each decoding step $t$ is computed by taking the dot product of the hidden state of the decoder at step $t$ and the projected encoder state for each word in the utterance, normalized using the softmax function. The output of a decoding step $t$ is a probability distribution over all the logical predicates. The distribution is modeled as a dot product between a projection of the overall decoding state at time $t$, and the embedding for each logical predicate, normalized using the softmax function. The overall decoding state at time $t$ is the concatenation of the hidden state of the decoder at time $t$ and the context vector at time $t$. Decoding of a logical form given an utterance is done as a greedy search over all possible logical forms to output.
\paragraph{Evaluation Metric}
Our evaluation metric is parsing accuracy. We define parsing accuracy as the ratio of the correctly parsed utterances in the test set. An utterance is parsed correctly if its generated logical form matches the annotated logical form exactly. E.g. \texttt{(field (relation parent) (person john))} is a correct parse of ``parents of John". We report accuracy percentage in Tables \ref{tab:results-extension} and \ref{tab:results-transfer}.
\subsection{Discussion}
We design two different scenarios for one-shot parsing evaluation, namely the \emph{extension} and the \emph{transfer} scenarios. We define and discuss the results of each scenario in the following sections.  

\paragraph{Extension}
This scenario tests one-shot parsing of new rules within the same domain that the model is trained on. In this case, we hold out several rules and their corresponding utterance-logical form pairs for evaluation and train on the remaining ones.

The results of this experiment are presented in Table~\ref{tab:results-extension}. Each column indicates the evaluation results on each domain.
The \emph{full} scores refer to the overall percentage of correct parses on the entire test data, the \emph{d=2} scores refer to the percentage of correct parses on examples with logical form depth 2 and the \emph{d=3} scores refer to the percentage of correct parses on examples with logical form depth 3.

As it can be seen in the table, our model is able to consistently improve the sequence-to-sequence baseline on all the domains by a significant margin of $68.8\%$.

\paragraph{Transfer}
This scenario tests one-shot parsing of new rules in a domain different from the one that the model is trained on. 
For example, in the \emph{transfer} scenario for domain \emph{person} the model is trained on examples from the other domains \emph{restaurant}, \emph{event}, \emph{course}, \emph{animal}, and \emph{vehicle} and tested on samples from domain \emph{person}. This is harder compared to the extension setup since it requires generalization to a completely new domain.

The results of this section are reported in Table~\ref{tab:results-transfer}. As it can be seen, the performance is improved by up to $36.6\%$ compared to the baseline model.

In order to have a better understanding of the model and indicate why it is performing comparable to the baseline on some of the domains setting, we carry out a model analysis in the next section.

\paragraph{Model Analysis}
We add two variations of our model ORACLE-DISCRIM and PRETRAIN-ENC for the transfer scenario to see the performance improvements gained from replacing different components of our model with an oracle/near oracle. This identifies potential bottlenecks of the model. 

In ORACLE-DISCRIM, we load a trained model from LOOKUPADAPT but replace the discriminator with an oracle during evaluation. As the numbers in row 3 of Table~\ref{tab:results-transfer} show, using an oracle discriminator improves the model performance by more than 10\% in the \emph{vehicle} domain and \emph{course} domain, and by smaller amounts in the other domains. On one hand, this suggests that the discriminator has room for improvement. On the other hand, the numbers suggest that there is still a much larger room for improvement for the encoder, aligner, and look-up components. Our hypothesis is that the encoder components are likely the main bottleneck because during evaluation in the \emph{transfer} scenario they have to produce good representations for utterances and logical forms very different from the ones which they have seen during training. We found evidence supporting this hypothesis in the PRETRAIN-ENC variation.

In PRETRAIN-ENC, we pretrain our model on all the domains, which ensures that the two encoders see all domains. We then use the encoder parameters of this all-domain pretrained model to initialize the encoders in the experiments for the \emph{transfer} scenario. We then fix the encoders and train only the aligner, discriminator, and look-up parameters. This results in near perfect generalization to the test domain by the aligner, discriminator, and look-up components, which are trained only on the other domains. This suggests that the encoder is the main bottleneck of our model since using a near-oracle version boosts its performance to perfect. The results for this variation is shown in row 4 of Table ~\ref{tab:results-transfer}. 

\begin{table*}[h]
\centering \small
\resizebox{1.5\columnwidth}{!}{%
\begin{tabular}{cc|c|c|c|c|c|c|}
\cline{3-8}
& & \multicolumn{6}{|c|}{\textbf{extension}}\\
\cline{3-8}
& & person & restaurant & event & course & animal & vehicle\\
\hline
\multicolumn{1}{|c}{\multirow{3}{*}{Sequence-To-Sequence}}
    & full & 20.1 & 25.1 & 26.1&  19.9 & 25.0 & 18.8\\
\multicolumn{1}{ |c  }{}  
    & d=2& 25.0& 18.1 & 20.0& 31.8& 26.5& 25.9\\
\multicolumn{1}{ |c  }{}  
    & d=3& 13.9& 37.3& 42.4& 0.0& 21.5& 5.8\\
\hline
\multicolumn{1}{|c}{\multirow{3}{*}{LOOKUPADAPT}}
    & full & \bf 45.9 & \bf 61.7 & \bf 85.5 & \bf 79.2& \bf 63.6& \bf 87.6\\
\multicolumn{1}{ |c  }{}  
    & d=2& 56.7& 85.4& 94.5& 86.7& 70.1& 93.6\\
\multicolumn{1}{ |c  }{}  
    & d=3& 27.8& 20.0& 61.7&  66.7&  46.5&  76.5\\
\hline
\end{tabular}
}
\caption{Test Accuracy on the \textbf{extension} Dataset.}
\label{tab:results-extension}
\end{table*}

\begin{table*}[h]
\centering \small
\resizebox{1.5\columnwidth}{!}{%
\begin{tabular}{cc|c|c|c|c|c|c|}
 \cline{3-8}
 & & \multicolumn{6}{|c|}{\textbf{transfer}}\\
 \cline{3-8}
 & & person & restaurant & event & course & animal & vehicle\\
 \hline
\multicolumn{1}{|c}{\multirow{3}{*}{Sequence-To-Sequence}}
    & full &  5.2 & 29.2 & 5.2&  \bf 19.9 & \bf 35.5& \bf 12.5\\
\multicolumn{1}{ |c  }{}  
    & d=2& 6.4&  41.7& 4.9&  27.5&  44.9& 15.6\\
\multicolumn{1}{ |c  }{}  
    & d=3& 3.1&  8.4&  5.7&  6.0& 13.8& 6.3\\
 \hline
\multicolumn{1}{|c}{\multirow{3}{*}{LOOKUPADAPT}}
    & full &  \bf 41.8& \bf  43.8&  \bf 8.4&  16.7&  28.3&  11.5\\
\multicolumn{1}{ |c  }{}  
    & d=2&  52.5& 53.4&  13.2&  24.2&  37.3&  14.1\\
\multicolumn{1}{ |c  }{}  
    & d=3&  21.4&  27.8& 0.0&  3.0&  6.9&  6.3\\
\hline \hline
\multicolumn{1}{|c}{\multirow{3}{*}{ORACLE-DISCRIM}}
    & full  &  46.9 &   45.8& 12.5  & 29.2&  33.3&  25.0\\
\multicolumn{1}{ |c  }{}  
    & d=2   & 55.6  &   56.7& 18.0  & 41.8&  44.8& 29.7\\
\multicolumn{1}{ |c  }{}  
    & d=3   & 30.3  &  27.8 & 2.9   &  5.9&  6.9&  15.6\\
\hline
\multicolumn{1}{|c}{\multirow{3}{*}{PRETRAIN-ENC}}
    & full  & 98.0& 100.0& 100.0& 99.0&  100.0&96.9\\
\multicolumn{1}{ |c  }{}  
    & d=2 & 100.0& 100.0& 100.0& 100.0& 100.0&  100.0\\
\multicolumn{1}{ |c  }{}  
    & d=3 & 94.0& 100.0& 100.0& 97.1& 100.0& 90.1\\
\hline
\end{tabular}
}
\caption{Test Accuracy on the \textbf{transfer} Dataset.}
\label{tab:results-transfer}
\end{table*}
\section{Related Work}
Semantic parsing is the task of mapping utterances to a formal representation of their meaning. Researchers have used grammar-based methods as well as machine learning-based methods to address this problem. Grammar-based parsers work by having a 
set of grammar rules that are either learned or hand-written, an algorithm for generating a set of candidate logical forms by recursive application of the grammar rules, and a criterion for picking the best candidate logical form within that set \cite{Liang2015Bringing, zettlemoyer2005learning, zettlemoyer2007online}. However, they are brittle to the flexibility of language. To improve this limitation, supervised sequence-based neural semantic parsers have been proposed \cite{nerualsemanticparsing}. \citet{herzig2017neural} improved the performance of neural semantic parsers by training over multiple knowledge bases and providing the domain encoding at decoding time. In addition to supervised learning, reinforcement learning methods for neural semantic parsing have been explored in \cite{zhong2017seq2sql}. 

Retrieve-and-edit style semantic parsing is gaining popularity. \citet{re} proposed a retrieve and edit framework that can efficiently learn to embed utterances in a task-dependent way for easy editing. Our work differs in that we perform hierarchical retrievals and edits, and that we evaluate on cross-domain data and focus on one-shot semantic parsing. It is worth noting that retrieve-and-edit as a general framework is not limited to semantic parsing and is applicable to other areas such as sentence generation. \cite{guu2018generating} and machine translation \citet{gu2018search}.

Another line of research maps semantic parsing under cross-domain setting to a domain adaptation problem \cite{su2017cross}. In their work, the model is trained on a certain domain and then fine tuned to parse data from another domain. This is in essence different from our work in that we do not adapt the model, rather we adapt seen samples to form parses of new samples. Moreover, We do not fine-tune any part of the model in the new domain and focus on one-shot semantic parsing.

Most of these models need many data-points to train. Therefore, there has been recent attempts at zero-shot semantic parsing. \citet{dadashkarimi2018zero} proposed a transfer learning approach where a domain label is predicted first and then the parse.  \citet{ferreira2015zero} and \citet{herzig2018decoupling} proposed slot-filling methods for semantic parsing based on general word embeddings.
\citet{bapna2017towards} focuses on zero-shot frame semantic parsing by leveraging the description of slots to be filled.

\section{Conclusion and Future Work}
As speech recognition technologies mature, more computing devices support spoken instructions via a conversational agent. However, most agents do not adapt to the phrasing and interest of a specific end user. It has recently been shown that new functionalities can be added to an agent from user instruction \cite{LIA, LIAv2}. However, the user instruction only provides one instance of a general instruction template and the agent is challenged to generalize to variations of the instance given during instruction. We define the \emph{one-shot parsing} task for measuring a semantic parser's ability to generalize to new instances of user-taught commands from only one example. We propose a new semantic parser architecture that learns a general strategy of retrieving seen utterances similar to an unseen utterance and adapting the logical forms of seen utterances to fit the unseen utterance. Our results show an improvement of up to 68.8\% on one-shot parsing under two different evaluation settings compared to the baselines. We found that the BiLSTM encoders are likely bottlenecks for the model. Some future directions include exploring the effects of contents in the memory, automating memory extraction from dataset, and improving the encoder.
\section*{Acknowledgments} This work was supported by AFOSR under grant FA95501710218, NSF under grant IIS1814472, and a Faculty award from J. P. Morgan.The authors would like to sincerely thank Bishan Yang for the initial discussions and ideas related to model architecture, and to Kathryn Mazaitis for the brainstorming sessions on the limitations of the model and future directions. 


\clearpage
\newpage

\appendix

\section{Sample Grammar and Data Generation Grammar}
\label{app:scfg}
A sample SCFG is given in Table~\ref{tab:example_scfg_person} to facilitate the definition of $domain$. 

This sample SCFG is not to be confused with the grammar used to generate our dataset. In comparison, the SCFG for data generation contains a larger number of entities, relations, and operations than the sample SCFG. Selected rules from the person domain and restaurant domain are provided below in Table~\ref{tab:person_grammar} and Table~\ref{tab:restaurant_grammar}.
\label{sec:rules}

\section{Contents of the Memory}
\label{app:memory}
In this paper, we assume the memory is provided to the model by a human expert. We picked the strategy of selecting one example for each grammar rule $g$ in the synchronous context free grammar $G$ that generated the data. It makes the formulation of maximum conditional likelihood learning straightforward, because this strategy produces a memory such that for any given utterance there is a unique sequence of \emph{look-up} and \emph{adapt} actions that produce the correct logical form. Recall that an example of a grammar rule $g$ is an \textit{utterance}-\textit{logical form} pair generated using $g$ and other rules of $G$. In particular, we choose an example for $g$ such that the top level predicate of the example's logical form matches the top level predicate of the target of $g$. For example, we will choose to put \par
$\langle$ John 's parents, \texttt{(field parent john)} $\rangle$ in memory for the grammar rule

\begin{align*}
    g_1 :=&\textbf{FIELD}  \to \\ &\langle\textbf{PSN$_2$}\ \text{'s}\ \textbf{PSN\_REL$_1$} ,\\
    & \texttt{(field}\  \textbf{PSN\_REL$_1$}\ \textbf{PSN$_2$}\texttt{)} \rangle
\end{align*}

because \texttt{field} is the root predicate of both the example logical form \texttt{(field parent john)} and the target of the rule \texttt{(field \textbf{PSN\_REL$_1$} \textbf{PSN$_2$})}. Here's another example. For a terminal rule like

\begin{equation*}
    g_2 :=\textbf{PSN}  \to \langle John,\ \texttt{(person john)} \rangle
\end{equation*}

we put the pair $\langle$ John,\ \texttt{(person john)} $\rangle$ in memory, because the top level predicate of the logical form (\texttt{person}) matches that of $g_2$'s target (\texttt{person}). We would not put for instance $\langle$ John 's children, \texttt{(field children john)}$\rangle$ because the top level predicate of the logical form (\texttt{field}) does not match that of $g_2$'s target (\texttt{person}). \par 

We do note that the requirement of manually selected examples to put in memory is a big limitation of the current approach. In future work, we wish to automate the building of memory.\par 

\section{The trace of an example run of Look-up and Adapt}
A trace of an example run of the \textsc{LookUpAndAdapt} algorithm is given in Table~\ref{tab:trace}. The utterance of this run is based on the sample SCFG grammar given in Table~\ref{tab:example_scfg_person}. Given a memory with 3 examples and a new utterance ``John's Parents'', the \textsc{LookUpAndAdapt} algorithm will look-up an example from the memory that matches the attended parts of the new utterance and recursively adapt the logical form of the retrieved example to produce a correct logical form that fits the new utterance.

\section{Statistics of the generated data}
\label{sec:data}
We provide a breakdown of the number of examples used in training and evaluation for the six domains of discourse at different depths in Table \ref{tab:dataStats}. The definition for the different groups of examples and their use is explained in \S~\ref{sec:dataset_generation} for dataset generation. 

\begin{table*}[t!]
\centering
\begin{tabular}{|llll|}
\hline \hline
Left-hand-side & & Source(Utterance) & Target(Logical Form)\\
\hline
\hline
\textbf{PSN} & $\to$ & John  & \texttt{john}\\
\hline 
\textbf{PSN} & $\to$ & Mary  & \texttt{mary}\\
\hline
\textbf{PSN\_REL} & $\to$ & parents & \texttt{parent}\\
\hline
\textbf{PSN\_REL} & $\to$ & children & \texttt{child}\\
\hline
\textbf{FIELD} & $\to$  &\textbf{PSN\_REL$_1$} of \textbf{PSN$_2$} & \texttt{(field} \textbf{PSN\_REL$_1$} \textbf{PSN$_2$}\texttt{)}\\
\hline
\textbf{FIELD} & $\to$  & \textbf{PSN$_2$} 's \textbf{PSN\_REL$_1$}  & \texttt{(field} \textbf{PSN\_REL$_1$} \textbf{PSN$_2$}\texttt{)}\\
\hline
\textbf{S} & $\to$ & \textbf{PSN$_1$} & \textbf{PSN$_1$}\\
\hline 
\textbf{S} & $\to$ & \textbf{PSN\_REL$_1$} & \textbf{PSN\_REL$_1$}\\
\hline 
\textbf{S} & $\to$ & \textbf{FIELD$_1$} & \textbf{FIELD$_1$}\\
\hline
\hline
\end{tabular}
\caption{Sample SCFG $G_{sample}$. The subscripts indicate linking of terminals/nonterminals in source and target. A SCFG rule can only be applied to linked terminals/nonterminals together.}
\label{tab:example_scfg_person}
\end{table*}
\begin{table*}[t!]
\small
\centering
\begin{tabular}{cc|c|c|c|c|c|c|}
\cline{3-8}
& & \multicolumn{6}{|c|}{\textbf{domain of discourse}}\\
\cline{3-8}
& & person & restaurant & event & course & animal & vehicle\\
\hline
\multicolumn{1}{ |c  }{\multirow{3}{*}{$D_{old}$}}
    & full  & 737   & 810   & 911   &  960  & 801  & 729\\
\multicolumn{1}{ |c  }{}  
    & d=2   & 417   & 490   & 591   & 640   & 481  & 409\\
\multicolumn{1}{ |c  }{}  
    & d=3   & 320   & 320   & 320   & 320   & 320  & 320\\
\hline
\multicolumn{1}{ |c  }{\multirow{3}{*}{$E_{olddev}$}}
    & full  & 96    & 96    & 96    & 96    & 96    & 96\\
\multicolumn{1}{ |c  }{}  
    & d=2   & 65    & 68    & 67    & 66    & 61    & 64\\
\multicolumn{1}{ |c  }{}  
    & d=3   & 31    & 28    & 29    & 30    & 35    & 32\\
\hline
\multicolumn{1}{ |c  }{\multirow{3}{*}{$E_{oldtest}$}}
    & full  & 96    & 96    & 96    & 96    & 96    & 96\\
\multicolumn{1}{ |c  }{}  
    & d=2   & 63    & 60    & 61    & 62    & 67    & 64\\
\multicolumn{1}{ |c  }{}  
    & d=3   & 33    & 36    & 35    & 34    & 29    & 32\\
\hline
\multicolumn{1}{ |c  }{\multirow{3}{*}{$E_{newdev}$}}
    & full  & 96    & 96    & 96   & 96    & 96    & 96\\
\multicolumn{1}{ |c  }{}  
    & d=2   & 68    & 67    & 58   & 68    & 60    & 66\\
\multicolumn{1}{ |c  }{}  
    & d=3   & 28    & 29    & 38   & 28    & 36    & 30\\
\hline
\multicolumn{1}{ |c  }{\multirow{3}{*}{$E_{newtest}$}}
    & full  & 96    & 96    & 96    & 96   & 96   & 96\\
\multicolumn{1}{ |c  }{}  
    & d=2   & 60    & 61    & 70    & 60   & 68   & 62\\
\multicolumn{1}{ |c  }{}  
    & d=3   & 36    & 35    & 26    & 36   & 28   & 34\\
\hline
\multicolumn{1}{ |c  }{\multirow{3}{*}{$M_{old}$}}
    & full  & 23    & 16    & 16    & 16    & 16    & 15\\
\multicolumn{1}{ |c  }{}  
    & d=1   & 20    & 13    & 13    & 13    & 13    & 12\\
\multicolumn{1}{ |c  }{}  
    & d=2   & 2     & 2     & 2     & 2     & 2     & 2\\
\multicolumn{1}{ |c  }{}  
    & d=3   & 1     & 1     & 1     & 1     & 1     & 1\\
\hline
\multicolumn{1}{ |c  }{\multirow{3}{*}{$M_{new}$}}
    & full  & 27    & 20    & 20    & 20    & 20    & 19\\
\multicolumn{1}{ |c  }{}  
    & d=1   & 24    & 17    & 17    & 17    & 17    & 16\\
\multicolumn{1}{ |c  }{}  
    & d=2   & 2     & 2     & 2     & 2     & 2     & 2\\
\multicolumn{1}{ |c  }{}  
    & d=3   & 1     & 1     & 1     & 1     & 1     & 1\\
\hline
\end{tabular}
\caption{Dataset Statistics: sizes of each component with depth breakdown. Please refer to \S~\ref{sec:dataset_generation} for the precise definition of each row of the table. In the \emph{extension} scenario, for each domain of discourse $d$, the model is trained on examples from $D_{old,d}$ using $M_{old,d}$ as memory and is evaluated on examples from $E_{newdev,d}$ and $E_{newtest,d}$ using $M_{new,d}$ as memory. In the \emph{transfer} scenario, for each domain of discourse $d$, the model is trained on $\cup_{d'\neq d} D_{old,d'}$ using $\cup_{d'\neq d} M_{old,d'}$ as memory, and is evaluated on $E_{olddev,d}$ and $E_{oldtest,d}$ using $\cup_{\forall d'} M_{old,d'}$ as memory.}
\label{tab:dataStats}
\end{table*}
\begin{table*}[t!]
\centering
\begin{tabular}{|llll|}
\hline \hline
Left-hand-side & & Source(Utterance) & Target(Logical Form)\\
\hline
\hline
\textbf{PSN} & $\to$ & john  & \texttt{(person john)}\\
\hline 
\textbf{PSN} & $\to$ & mary  & \texttt{(person mary)}\\
\hline
\textbf{PSN} & $\to$ & colleagues  & \texttt{(person colleague)}\\
\hline
\textbf{PSN} & $\to$ & professors  & \texttt{(person professor)}\\
\hline
\textbf{PSN\_REL} & $\to$ & parents & \texttt{(relation parent)}\\
\hline
\textbf{PSN\_REL} & $\to$ & children & \texttt{(relation child)}\\
\hline
\textbf{PSN\_REL} & $\to$ & hometown & \texttt{(relation hometown)}\\
\hline
\textbf{PSN\_REL} & $\to$ & salary & \texttt{(relation salary)}\\
\hline
\textbf{FIELD} & $\to$  &\textbf{PSN\_REL$_1$} of \textbf{PSN$_2$} & \texttt{(field} \textbf{PSN\_REL$_1$} \textbf{PSN$_2$}\texttt{)}\\
\hline
\textbf{FIELD} & $\to$  & \textbf{PSN$_2$} 's \textbf{PSN\_REL$_1$}  & \texttt{(field} \textbf{PSN\_REL$_1$} \textbf{PSN$_2$}\texttt{)}\\
\hline
\textbf{CMD} & $\to$  & set \textbf{FIELD$_1$} to \textbf{VALUE$_2$}  & \texttt{(set} \textbf{FIELD$_1$} \textbf{VALUE$_2$}\texttt{)}\\
\hline
\textbf{VALUE} & $\to$ & \textbf{FIELD$_1$} & \textbf{FIELD$_1$}\\
\hline
\textbf{VALUE} & $\to$ & \textbf{PSN$_1$} & \textbf{PSN$_1$}\\
\hline
\textbf{S} & $\to$ & \textbf{FIELD$_1$} & \textbf{FIELD$_1$}\\
\hline
\textbf{S} & $\to$ & \textbf{CMD$_1$} & \textbf{CMD$_1$}\\
\hline
\hline
\end{tabular}
\caption{This is a selected subset of the grammar used to generate data for the \emph{person} domain. Rules not shown here include those for introducing flexibility in language and rules for additional entities and relations.}
\label{tab:person_grammar}
\end{table*}

\begin{table*}[t!]
\centering
\begin{tabular}{|llll|}
\hline \hline
Left-hand-side & & Source(Utterance) & Target(Logical Form)\\
\hline
\hline
\textbf{RST} & $\to$ & chinese restaurant  & \texttt{(restaurant chinese)}\\
\hline 
\textbf{RST} & $\to$ & italian restaurant  & \texttt{(restaurant italian)}\\
\hline
\textbf{RST} & $\to$ & french restaurant  & \texttt{(restaurant french)}\\
\hline
\textbf{RST} & $\to$ & german restaurant  & \texttt{(restaurant german)}\\
\hline
\textbf{RST\_REL} & $\to$ & address & \texttt{(relation address)}\\
\hline
\textbf{RST\_REL} & $\to$ & reviews & \texttt{(relation reviews)}\\
\hline
\textbf{RST\_REL} & $\to$ & dishes & \texttt{(relation dishes)}\\
\hline
\textbf{RST\_REL} & $\to$ & price & \texttt{(relation price)}\\
\hline
\textbf{FIELD} & $\to$  &\textbf{RST\_REL$_1$} of \textbf{RST$_2$} & \texttt{(field} \textbf{RST\_REL$_1$} \textbf{RST$_2$}\texttt{)}\\
\hline
\textbf{FIELD} & $\to$  & \textbf{PSN$_2$} 's \textbf{RST\_REL$_1$}  & \texttt{(field} \textbf{RST\_REL$_1$} \textbf{RST$_2$}\texttt{)}\\
\hline
\textbf{CMD} & $\to$  & set \textbf{FIELD$_1$} to \textbf{VALUE$_2$}  & \texttt{(set} \textbf{FIELD$_1$} \textbf{VALUE$_2$}\texttt{)}\\
\hline
\textbf{VALUE} & $\to$ & \textbf{FIELD$_1$} & \textbf{FIELD$_1$}\\
\hline
\textbf{VALUE} & $\to$ & \textbf{RST$_1$} & \textbf{RST$_1$}\\
\hline
\textbf{S} & $\to$ & \textbf{FIELD$_1$} & \textbf{FIELD$_1$}\\
\hline
\textbf{S} & $\to$ & \textbf{CMD$_1$} & \textbf{CMD$_1$}\\
\hline
\hline
\end{tabular}
\caption{This is a selected subset of the grammar used to generate data for the \emph{restaurant} domain. Rules not shown here include those for introducing flexibility in language and rules for additional entities and relations.}
\label{tab:restaurant_grammar}
\end{table*}
\begin{table*}[h!]
\centering
\footnotesize
\begin{tabular}{|@{\hskip2pt}l@{\hskip3pt}|@{\hskip3pt}l@{\hskip3pt}|@{\hskip3pt}c@{\hskip3pt}|@{\hskip3pt}c@{\hskip2pt}|}
\hline \hline
 Func & Description  &
 \shortstack{New utterance and\\
            current logical form } &
 Memory\\
\hline \hline

    \textsc{LkUpNAdpt}&
    \shortstack[l]{Initial call always has\\
                    uniform attention.} &
    \shortstack{\textbf{[John 's parents]}\\
                None} &
    \shortstack{$\langle$John, \texttt{john}$\rangle$\\ 
                $\langle$Mary, \texttt{mary}$\rangle$\\
                $\langle$Mary 's parents ,\texttt{(field parent mary)}$\rangle$}\\
\hline
    \textsc{LookUp}&
    \shortstack[l]{Retrieves entry $\star$ from\\
                    memory by utterance\\
                    similarity.} &
    \shortstack{\textbf{[John 's parents]}\\
                \texttt{(field parent mary)}} &
    \shortstack{$\langle$John, \texttt{john}$\rangle$\\ 
                $\langle$Mary, \texttt{mary}$\rangle$\\
                $\star$ $\langle$Mary 's parents ,\texttt{(field parent mary)}$\rangle$}\\
\hline
    ::\textsc{Adapt}&
    \shortstack[l]{Adapt first child.\\
            \texttt{parent} is aligned to\\
            \textbf{parents}. YES, they\\
            match, keep \texttt{parent}.\\
            There's no children. \\
            Return.} &
    \shortstack{John 's \textbf{[parents]}\\
                \texttt{(field [parent] mary)}} &
    \shortstack{$\langle$John, \texttt{john}$\rangle$\\ 
                $\langle$Mary, \texttt{mary}$\rangle$\\
                 $\langle$Mary 's parents ,\texttt{(field parent mary)}$\rangle$}\\
\hline
    ::\textsc{Adapt}&
    \shortstack[l]{Adapt second child.\\
            \texttt{mary} is aligned to\\
            \textbf{John}. NO, they\\
            don't match.\\
            Find replacement and\\
            return it.} &
    \shortstack{\textbf{[John]} 's parents\\
                \texttt{(field parent [mary])}} &
    \shortstack{$\langle$John, \texttt{john}$\rangle$\\ 
                $\langle$Mary, \texttt{mary}$\rangle$\\
                 $\langle$Mary 's parents ,\texttt{(field parent mary)}$\rangle$}\\
\hline
    ::::\textsc{LkUpNAdpt}&
    \shortstack[l]{Find replacement\\
                    for \texttt{mary}.} &
    \shortstack{\textbf{[John]} 's parents\\
                \texttt{(field parent mary)}} &
    \shortstack{$\langle$John, \texttt{john}$\rangle$\\ 
                $\langle$Mary, \texttt{mary}$\rangle$\\
                $\langle$Mary 's parents ,\texttt{(field parent mary)}$\rangle$}\\
\hline
    ::::\textsc{LookUp}&
    \shortstack[l]{Retrieves entry $\star$ from\\
                    memory by utterance\\
                    similarity.} &
    \shortstack{\textbf{[John]} 's parents\\
                \texttt{(field parent mary)}} &
    \shortstack{$\star$ $\langle$John, \texttt{john}$\rangle$\\ 
                $\langle$Mary, \texttt{mary}$\rangle$\\
                $\langle$Mary 's parents ,\texttt{(field parent mary)}$\rangle$}\\
\hline
    ::::\textsc{LkUpNAdpt}&
    \shortstack[l]{There's no children. \\
                Return \texttt{john}} &
    \shortstack{\textbf{[John]} 's parents\\
                \texttt{(field parent mary)}} &
    \shortstack{$\langle$John, \texttt{john}$\rangle$\\ 
                $\langle$Mary, \texttt{mary}$\rangle$\\
                $\langle$Mary 's parents ,\texttt{(field parent mary)}$\rangle$}\\
\hline
    \textsc{LkUpNAdpt}&
    \shortstack[l]{Replace \texttt{mary} with\\
                   \texttt{john}. Exhausted\\
                   all children, return\\
                   the current logical \\
                   form.} &
    \shortstack{\textbf{[John 's parents]}\\
                \texttt{(field parent john)}} &
    \shortstack{$\langle$John, \texttt{john}$\rangle$\\ 
                $\langle$Mary, \texttt{mary}$\rangle$\\
                $\langle$Mary 's parents ,\texttt{(field parent mary)}$\rangle$}\\
\hline
\hline
\end{tabular}
\caption{Trace of a sample run of the algorithm with sample memory, utterance, and attention. The attention weights (argument $\bm{w}'$ to \textsc{LookUpAndAdapt} and variable $\bm{w}$ in \textsc{Adapt}) are visualized as brackets. Bracketed words in the \emph{utterance} receive more attention, un-bracketed words receive less attention. For \textsc{Adapt} calls, bracketed words in the \emph{logical form} indicate the current subtree of the whole logical form tree that the algorithm is looking at (the argument \texttt{T} to \textsc{Adapt}). Colons are used to indicate the depth of the call stack.}
\label{tab:trace}
\end{table*}
\end{document}